# DFVS: Deep Flow Guided Scene Agnostic Image Based Visual Servoing

Y V S Harish[1], Harit Pandya[2], Ayush Gaud[3], Shreya Terupally[1], Sai Shankar[1] and K. Madhava Krishna[1]

*Abstract*— Existing deep learning based visual servoing approaches regress the relative camera pose between a pair of images. Therefore, they require a huge amount of training data and sometimes fine-tuning for adaptation to a novel scene. Furthermore, current approaches do not consider underlying geometry of the scene and rely on direct estimation of camera pose. Thus, inaccuracies in prediction of the camera pose, especially for distant goals, lead to a degradation in the servoing performance. In this paper, we propose a two-fold solution: (i) We consider optical flow as our visual features, which are predicted using a deep neural network. (ii) These flow features are then systematically integrated with depth estimates provided by another neural network using interaction matrix. We further present an extensive benchmark in a photo-realistic 3D simulation across diverse scenes to study the convergence and generalisation of visual servoing approaches. We show convergence for over 3m and 40 degrees while maintaining precise positioning of under 2cm and 1 degree on our challenging benchmark where the existing approaches that are unable to converge for majority of scenarios for over 1.5m and 20 degrees. Furthermore, we also evaluate our approach for a real scenario on an aerial robot. Our approach generalizes to novel scenarios producing precise and robust servoing performance for 6 degrees of freedom positioning tasks with even large camera transformations without any retraining or fine-tuning.

## I. INTRODUCTION

Visual servoing addresses the problem of attaining a desired pose with respect to a given environment using image measurements from a vision sensor. Classical visual servoing approaches extract a set of hand-crafted features from the images. Pose based visual servoing (PBVS) approaches use these visual features to estimate the camera pose directly in Cartesian space from a given image. The controller then guides the robotic system in the direction that minimizes the difference in pose between current and desired image pair directly in 3D space. In contrast, image based visual servoing (IBVS) approaches control the robot by minimizing the feature error explicitly in the image space [1]. It can be observed that the pose based visual servoing controllers attain the desired pose without getting stuck at local minima. They, however, are sensitive to camera calibration errors and pose estimation errors [2]. On the contrary, image based visual servoing approaches are robust to calibration and depth errors but could lead to a local minima. Classical PBVS and IBVS approaches, both rely on reliable matching of hand-crafted features, thus inaccuracies while obtaining correspondences degrades the servoing performance. Direct visual servoing [3] approaches avoid the feature extraction step and operate directly on image measurements. This helps in achieving higher precision in goal reaching, but the trade-off is a smaller convergence basin. Another rigid requirement of classical visual servoing approaches is the knowledge of environment's depth. This is especially difficult to obtain on robotic systems using a monocular camera.

To circumvent the requirement for extracting and tracking hand-crafted features, Saxena et al. [4] presented a deep learning based visual servoing approach. Specifically, they employed a deep network to estimate the relative camera pose, from an image pair. A traditional PBVS controller is then used to minimize the relative pose between the current and the desire image. Their network was trained on publicly available Microsoft 7 scenes dataset [5] for estimating relative camera pose. Although trained on limited number of scenes, their network was able to generalise well on novel environments, however, the convergence basin was limited. Bateux et al. [6] presented a similar deep pose based visual servoing approach with a Siamese [7] based network architecture for estimating relative camera pose from an image pair. They further proposed extensive guidelines for training deep networks for the task of visual servoing. They used LabelMe database [8] which contains a diverse set of images with scene variations while using homography for obtaining viewpoint variations to make the network robust. The network was then trained to estimate the relative pose given a pair of images taken from these viewpoints, which helped in generalisation of the approach to different environments. Similarly, Yu et al. also present a Siamese style deep network for visual servoing [9], their network obtains a much higher sub-millimeter precision for the servoing task, however the network was trained only on a table-top scene with similar objects and therefore requires retraining for adjusting to novel environments.

Unlike the above approaches that estimate the relative camera pose and use a PBVS controller for achieving the desired pose, recent deep reinforcement learning based visual servoing approaches [10], [11], [12], [13] use neural controllers to maximize the rewards and therefore require minimal supervision. However, several of these approaches are specific to manipulation tasks and are trained only for scene with objects lying on a table. Furthermore, these approaches do not consider full 6 degrees of freedom (DOF) visual servoing. Sampedro et al. [14] showcased a similar deep reinforcement learning approach for an aerial robot for autonomous landing on a moving target, however they only report results for a single scene with a colored target. Zhu et al. [15] presents the results quite similar to ours on a wide range of environments. Although, they constrain the motion to a plane and their approach is showcased on similar environments, thus generalisation on novel environments is non-trivial and requires fine tuning.

[1]IIIT Hyderabad, India, harish.y@research.iiit.ac.in, shreya.reddy@students.iiit.ac.in, mkrishna@iiit.ac.in
[2]University of Lincoln, UK, hpandya@lincoln.ac.uk
[3]Mathworks, India, ayush.gaud@gmail.com

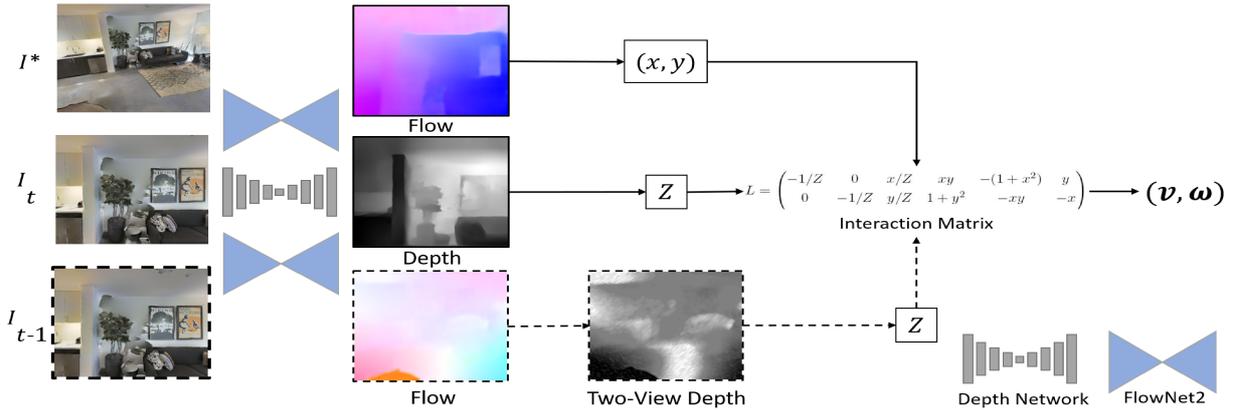

Fig. 1: Pipeline of the proposed approach. We employ Flownet-2 network to estimate the feature error between current and desired pose. To predict the scene's depth at the current pose, we propose two designs: (i) We employ a depth-network to estimate depth from a single view. (ii) Alternatively(dashed line), a scaled version of flow(calculated empirically) could also be used as disparity to achieve scene agnostic depth prediction. Finally, we combine the cues from depth and visual features using an interaction matrix similar to classical IBVS. This helps us in achieving a robust and precise visual servoing.

In this paper, we present a visual servoing approach based on deep learning for attaining a 6 DOF pose, which effectively generalises to novel environments. In contrast to existing approaches that directly attempt to estimate the relative pose between current and desired pose, we compute the dense correspondences using an optical flow network. These are combined with depth estimates from another network in a principled manner using interaction matrix. Such close integration of dense correspondences and with depth estimates of the scene allows the network in achieving a large convergence basin and more generalisation to novel environments. To benchmark against existing deep visual servoing approaches, we extensively evaluate them on numerous perception and control criteria in photo-realistic simulation environments. Our approach not only attains precise convergence but it also showcases larger convergence basin as compared to existing approaches.

Our contributions are summarized as follows:
- We propose a novel system consisting of deep neural networks that systematically integrates depth cues with flow features. Our approach exhibits precise positioning even for distant goals achieving a convergence uptill the 4m and 40° difference in pose (shown in section III-C), where existing approaches fails to converge. Our model also is able to generalise in diverse environments both indoors and outdoors effortlessly without any need for fine-tuning or retraining unlike other approaches [15].
- We present an extensive benchmark for deep servoing methods evaluating both perception and control performance on a variety of parameters such as, photometric error, camera poses error, trajectory length and convergence basin. We showcase state-of-the-art performance under this comprehensive evaluation in table I.
- We make our implementation and the pipeline to benchmark visual servoing approaches publicly available[1]. It consists of photo-realistic scenes classified into three categories as easy, medium and hard based on the scene texture and convergence basin. It will help in facilitating fair and quantitatively thorough comparison of various visual servoing schemes.

## II. Approach

Existing deep visual servoing approaches aim to directly estimate the relative camera pose between the current and the desired pose in a PBVS style, therefore these approaches require a data sample for every iteration. This amounts to a huge amount of training data which is difficult to obtain. So, Bateux et al. [6] trained their network using dataset generated on a planar scene from multiple viewpoints in a simulation environment. Moreover, by selecting the textures from a variety of backgrounds, generalization could be achieved over different environments. Although, due to the incorrect depth of the scene, it is non-trivial to attain large camera transformations. Saxena et al. [4] on the other hand, train their data by sampling image pairs from a given trajectory, as a result they trained their network on a limited dataset acquired from just seven environments. Yu et al. [9] trained their networks only on a single table-top setting.

To circumvent these strict training requirements, we base our approach on image based visual servoing. We employ a flow network [16] to estimate dense correspondences between current view and desired view. These visual features are easier to observe as compared to the relative pose estimates. Furthermore, to capture the geometry of the scene, we pivot our approach on depth cues similar to classical image based visual servoing. However, unlike classical IBVS, where a coarse depth was assumed, we are able to estimate the depth by using a deep network. In this work, we present two designs for obtaining the depth, in the first method we explicitly use a depth network [17] while our second design employs a scaled version of flow as a proxy for the depth. Our pipeline is illustrated in figure 1.

[1]Project page: https://harishyvs.github.io/FlowBasedIBVS/

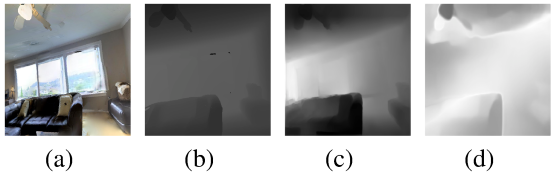

Fig. 2: (a) Given image, (b) True depth from the depth sensor, (c) single view depth from Depth Network and (d) two-view depth from flow network.

### A. Flow as visual features

Classical IBVS approaches use local appearance based features such as SURF [18], SIFT [19] to describe the scene and the feature error is computed by the difference between their locations in the current and the desired image $(s - s^*)$. This requires accurate tracking of the visual features. However, for low textured environments, the number of feature detections are less, this combined with large matching inaccuracies results in incorrect error estimates. Whereas, optical flow provides dense correspondences even in case of low textured environments. Therefore, in this work we use optical flow to estimate the feature error between the current and the desired image ($\mathscr{F}(I_t, I^*)$). For estimating the visual feature error, we use the FlowNet-2 [16], which is state-of-the-art in flow computation.

### B. Depth estimation

Recent deep visual servoing approaches directly estimate the camera pose without accounting for the geometry of the underlying scene. This increases the estimation errors for distant goal which in turn reduces the convergence basin. For instance, Bateux et al. [6] trained their network on planar scenes and report sharp decrease in converge ratio for distance greater 2.5 cm and angle greater than 15 degrees. Similarly, Saxena et al. [3] estimate the relative translation and rotation individually and thus require a scaling factor to fuse them.

Even for the classical image based visual servoing approaches, depth plays a crucial role in their convergence in 3D environments. In the absence of knowledge of true depth of the scene, classical IBVS approaches either approximate the true depth through a coarse depth [1], or estimate the depth online [20] by observing the robot's odometry. Which, again limits the convergence basin especially in scenarios involving large camera transforms. In this paper, we present two designs to accommodate depth cues for stable visual servoing described in the sections below.

*1) Single view depth estimation:* Our first design is based on directly estimating scene's depth from single image. We employ an encoder-decoder based convolutional neural network architecture proposed by Alhashim et al. [17]. Authors inspire their encoder design from the DenseNet-169 [21], whereas the decoder is composed of 3 up-sampling blocks, each consisting of an up-sampling layer which is then concatenated with respective layer from encoder, similar to the U-net [22] and is followed by 2 convolution layers.

*2) Two view depth estimation:* The depth network proposed in the previous sub-section can estimate the scene's depth from a single image. however to achieve scene agnostic depth prediction, as an alternate, we propose to estimate depth using an image pair. Specifically, magnitude of optical flow could be seen as an inverted scaled representation of underlying depth. We again rely on our flow network to compute the optical flow between current and previous image and use it as a proxy for the depth, which allows the network to generalise across a diverse set of indoor and outdoor scenes.

Figure 2 compares the depth of an indoor scene estimated from both the approaches as compared to the ground truth depth of the scene given by the depth sensor of our simulation engine. It can be easily seen that depth network does a much better job in predicting the depth but it was trained on indoor objects, however the two-view depth captures the geometry of the scene at a coarser level without any fine-tuning or refinement.

### C. Image based visual servoing

Classical visual servoing approaches consider sparse correspondences between the current and desired configuration and minimize them iteratively using gradient descent over feature error in image space and pseudo-inverse of image Jacobian. On the other hand, we use a deep network [16] for estimating the dense correspondences in the form of optical flow. Due to large dimensionality of the feature vector we employ levenberg-marquardt based gradient descent similar to collewet et al. [23], specifically our control law describing the camera velocity $v_c$ is given as:

$$v_c = -\lambda (L^T L + \mu\, diag(L^T L))^{-1} \mathscr{F}(I_t, I^*). \quad (1)$$

Where, the visual feature error is given by dense correspondences observed by the flow network. $L(z)$ is the interaction matrix as described in [1] and the depth $z$ is estimated by either the depth network $D(I_t)$ or the scaled flow $\alpha \mathscr{F}(I_t, I_{t-1})^{-1}$. The tunable parameters $\lambda$ and $\mu$ adjust the step size and direction of the descend respectively.

## III. EXPERIMENTS

The motivation of the current work is to achieve an off-the-shelf scene agnostic visual servoing. Several deep visual servoing approaches [10], [11], [12], [13] only focus on goal reaching task for manipulation and do not consider the pose in which object is reached. A few approaches such as [6], [9] do consider a 6 DOF goal reaching task, however their test-cases are limited to planar or near-planar scenarios like a table-top. Only [3] provide results for 6 DOF goal reaching task for non-planar scenes, however they do not show results in a photo-realistic environments. The difference between the environments used for the servoing task could easily be noticed in figure 3. The central focus in the experiments is to validate the ability to generalise on different environments. Therefore, we present a set of benchmarking tasks in simulation environment to be performed without fine-tuning the network. To showcase that our approach can be used as plug-and-play for various environments, we further present results with an aerial robot on an outdoor scenario.

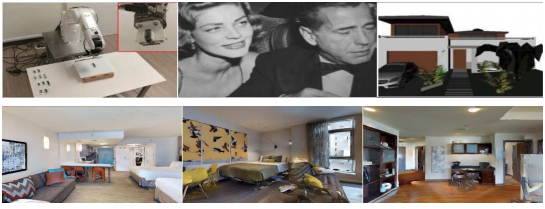

Fig. 3: The test-beds compared by existing deep visual servoing approaches (top) are either planar, near -planar(table-top) or synthetic. On the other hand, we propose a photo-realistic benchmark (bottom) for deep visual servoing approaches.

### A. Simulation results on Benchmark

The proposed simulation benchmark consists of 10 indoor photo-realistic environments from Habitat simulation engine [24]. We have selected these scenes such that they cover different textures and variable number of objects. Further for each scene we provide an initial pose and a desired pose. We classify these tasks in three categories easy (refer figure 4 row 1-3), medium (refer figure 4 row 4-7)and hard (refer figure 4 row 8-10).The categorization was done based on the complexity of the scenes namely the amount of texture present ,the extent of overlap and the rotational , translational complexities between initial and desired image.The first category, easy has quite a good number of distinctive objects in the scene and has more of just translational motion[around 1.4m] and small rotation[around 15°] ,the second category, medium has less number of objects(lesser texture) and also a decent change in rotation[near 20°] and translation[near 1.5m]. The last category, hard either has huge amount of rotation[>=30°] or translation[>=2m] or both thus having less overlap between initial and desired image. To evaluate the visual servoing approaches we propose following metrics capturing both perception as well as control aspects of servoing: final translation error, final rotation error, trajectory length and number of iterations . We report both quantitative table I and qualitative fig 4 results on the benchmark. It can be noticed that Saxena et al. [4], are able to converge on easy and medium scenes but they have difficulty on hard scenes. We also compared our approaches with photometric visual servoing, using true depth of the scene obtained from the depth sensor. It can be seen from the table I even with the knowledge of the correct depth of the scene, it is not able to converge in most of the environments. Quantitative results (table I) shows that when using depth predicted by our approaches (single-view as well as two-view), pose error after convergence is at par compared with ground truth depth for visual servoing tasks. The mean error with our approach after convergence is 0.025 cm and 1.167 degrees on 10 different scenes with initial mean pose error of 1.76 cm and 22.89 degrees. On the contrary, both [3] and [4] fail to converge on more than half the scenarios. The simulations done using true depth were stopped without achieving 100 % convergence [zero photometric error] since, initially we only wanted to know the the precision upto which we can go using true depth. Therefore, the results using true depth are close to the two proposed pipelines.

### B. Controller performance

For analysing the controller performance, we next present the results for a visual servoing trial. The initial pose and desired pose are given from figure 4, row 8. It can be observed from figure 5 that both our approaches (flow-depth based and depth-network based) are able to converge without any oscillations while [3] and [4] diverge. The photometric error steadily reduces. It can also be seen that flow-depth based approach takes longer to converge as compared to the depth-network based approach but has a much shorter trajectory. The velocity profile is bounded and gradually decreases to zero.

### C. Convergence study

In this experiment we compare existing approaches with ours for studying the convergence domain. We randomly select multiple scenes from habitat environment, vary the desired pose and evaluate how many times our approach was able to converge. Similar to [6], our flow-depth pipeline was evaluated by linearly increasing the distance between initial and desired image in each axis x,y,z by 0.4 meters till 4 meters, making 10 batches in total. We have selected the rotations in [x,y,z] into 3 set-points specifically [10°,10°,25°], [20°,20°,40°] ,[30°,30°,50°]. Thus, considering 3 different situations each having an increased translational variation in [x,y,z] (refer to the x axis in fig 6). Each batch of experiments had 16 environments randomly selected from Gibson dataset, upon which the convergence ratio was calculated, the initial position was fixed and the desired position was varied as mentioned. It can be seen from the figure 6 that our approach outperforms the existing approaches like [6], where the convergence ratio drastically drops to about 65 percent around the variance of 2.4cm , 12°in x and y and 1.2cm , 30°in z. Our pipeline showed greater convergence ratio of about 90 percent for the [20°,20°,40°] case which is close to the setting proposed by [6] even with a higher convergence basin till 4m. Our approach shows robust performance even for complex cases with [30°,30°,50°] rotational change [convergence for over 75 percent in most cases] where in [6] falls down to 40 percent at [20°,20°,50°]. Note that the criteria for convergence is final translation error less than 4 cm and rotation less than 1°.

### D. Real drone experiment

We finally validate the generalisation of our approach using Parrot Bebop-2 drone on outdoor scenario. We test our approach for a large camera transformation of [0.29, 0.39,1.27]m in translation and [-25.63°,-25.63°,-10.56°] in rotation between the initial and desired images, as shown in figure 7. It can be seen qualitatively that the robot is able to smoothly attain the desired pose precisely even with constant illumination variations in the outdoor scenario without need for fine-tuning or retraining. Note that due to erroneous odometry of the robot, we only present the qualitative results for this experiment.

| Initial Image | Desired Image | PhotoVS [3] | Saxena et al.[4] | True depth | Depth-net | Flow-depth |
|---|---|---|---|---|---|---|

Fig. 4: Qualitative results on the benchmark for 10 scenes. Given initial and desired images of various scenes from the benchmark, we compare 3 variants of our approach (true depth, network-depth and flow-depth) with Saxena et al. [4] and PhotoVS [3] visualize the error image between desired and resulting pose of the approach. While both [4] and [3] converges on all easy (row 1-3) and some medium scenes (row 4-7) they fail to converge on all hard scenes. Whereas, all the variants of our approach are able to converge on all test-cases. Grey areas mean that the difference between attained and desired image is zero and white portions indicate there is a slight non-overlap between them.

## IV. CONCLUSION

In this work, we have put forward a baseline comparison between all the present state of the art supervised visual servoing techniques and have quantified the effect of each method. We have compared our deep image based visual servoing technique with the existing frameworks. We have used a network to estimate the optical flow between the images and have used this as visual features. A major break through in the attainment of the desired pose can be found using the integration with depth estimates. We presented two methods for estimating depth under single and two-view settings. We Also presented an extensive benchmark to evaluate servoing. Our approach showcases precise servoing with large convergence basin for diverse environments and performs robustly without retraining or fine-tuning.

## V. ACKNOWLEDGEMENT

This work was supported by EPSRC under agreement EP/R02572X/1 NCNR.

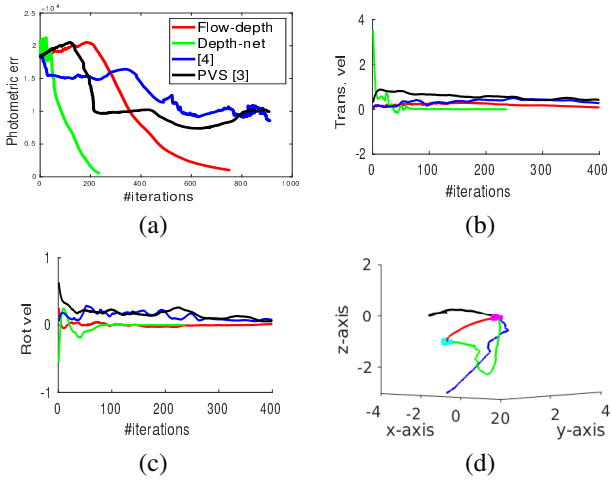

(a) (b)

(c) (d)

Fig. 5: 3D positioning task for an indoor scene: (a) Photometric feature error, (b) Translational velocity in m/s., (c) Rotational velocity in rad/s. and (d) Camera trajectory. Both the variants of our approach are able to converge, whereas existing approaches fail.

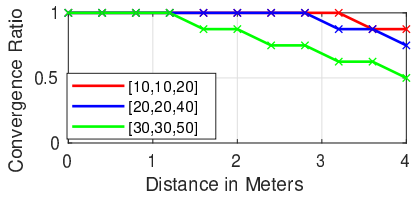

Fig. 6: Convergence study for medium and large camera transforms in simulation, the labels represent rotation in deg. [x,y,z]. Our approach shows larger convergence domain over existing approaches such as [6].

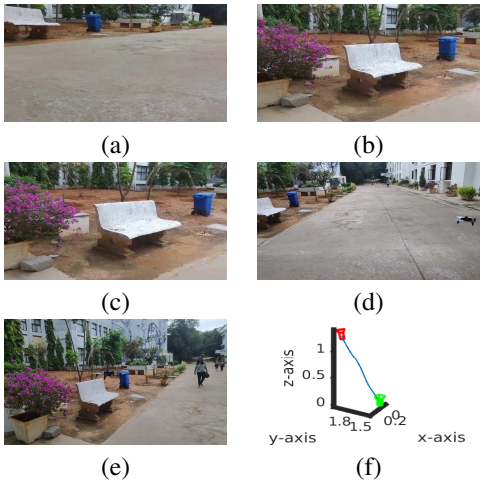

(a) (b)

(c) (d)

(e) (f)

Fig. 7: Outdoor positioning task: (a) Initial image as seen by the drone, (b) Desired final image (c) Attained image, (d) Initial drone position as seen by us (mid-right), (e) Final drone position as seen by us (center). and (f) Drone's Trajectory. Our approach is able to attain precise positioning for this task as well without any retraining or fine-tuning.

| Metric | I. err | [3] | [4] | T.depth | D.net | F.depth |
|---|---|---|---|---|---|---|
| T. err | 1.42 | 0.06 | 0.19 | 0.02 | 0.02 | 0.04 |
| R. err | 18.20 | 0.76 | 0.89 | 0.37 | 0.38 | 0.72 |
| Tj. len | - | 2.26 | 2.69 | 2.24 | 1.42 | 1.19 |
| Iter | - | 956 | 1486 | 764 | 233 | 1560 |
| T. err | 1.49 | 0.12 | 0.08 | 0.04 | 0.02 | 0.03 |
| R. err | 16.33 | 0.66 | 0.84 | 0.84 | 0.42 | 0.06 |
| Tj. len | - | 2.22 | 2.48 | 1.58 | 1.55 | 2.31 |
| Iter | - | 869 | 1687 | 150 | 143 | 871 |
| T. err | 1.41 | 0.28 | 0.13 | 0.03 | 0.02 | 0.04 |
| R. err | 18.07 | 3.23 | 6.44 | 8.72 | 8.87 | 8.39 |
| Tj. len | - | NC | 2.94 | 2.31 | 1.19 | 1.1 |
| Iter | - | NC | 1647 | 214 | 185 | 580 |
| T. err | 1.15 | 0.91 | 0.83 | 0.02 | 0.02 | 0.02 |
| R. err | 17.27 | 13.25 | 10.54 | 1.39 | 1.38 | 0.66 |
| Tj. len | - | NC | NC | 1.52 | 2.56 | 0.89 |
| Iter | - | NC | NC | 104 | 186 | 3831 |
| T. err | 1.64 | 0.33 | 0.04 | 0.03 | 0.03 | 0.03 |
| R. err | 30.96 | 3.57 | 1.39 | 0.85 | 0.83 | 0.89 |
| Tj. len | - | NC | 2.86 | 2.36 | 3.65 | 2.37 |
| Iter | - | NC | 1235 | 84 | 89 | 869 |
| T. err | 1.37 | 1.29 | 1.25 | 0.02 | 0.02 | 0.03 |
| R. err | 17.26 | 7.63 | 9.55 | 1.81 | 0.77 | 1.81 |
| Tj. len | - | NC | NC | 1.33 | 2.49 | 1.26 |
| Iter | - | NC | NC | 20 | 283 | 661 |
| T. err | 1.95 | 0.54 | 0.35 | 0.02 | 0.02 | 0.03 |
| R. err | 22.86 | 4.56 | 6.47 | 0.53 | 0.53 | 1.39 |
| Tj. len | - | NC | NC | 2.74 | 2.82 | 1.85 |
| Iter | - | NC | NC | 62 | 244 | 981 |
| T. err | 2.54 | 2.49 | 2.32 | 0.02 | 0.01 | 0.03 |
| R. err | 20.02 | 39.65 | 48.77 | 11.67 | 0.34 | 0.55 |
| Tj. len | - | NC | NC | 2.03 | 5.88 | 2.26 |
| Iter | - | NC | NC | 504 | 237 | 754 |
| T. err | 1.94 | 2.36 | 2.27 | 0.04 | 0.041 | 0.041 |
| R. err | 31.78 | 43.37 | 29.69 | 0.78 | 0.75 | 0.91 |
| Tj. len | - | NC | NC | 2.24 | 2.28 | 2.32 |
| Iter | - | NC | NC | 183 | 145 | 1124 |
| T. err | 2.43 | 2.27 | 1.36 | 0.01 | 0.01 | 0.02 |
| R. err | 36.05 | 53.246 | 29.24 | 0.28 | 0.16 | 0.49 |
| Tj. len | - | NC | NC | 2.52 | 2.67 | 3.52 |
| Iter | - | NC | NC | 314 | 114 | 1386 |

TABLE I: Quantitative results on benchmark: We compare our all the 3 variants of our approach True depth(T. depth), Depth network(D.net) and Flow depth(F.depth), with existing visual servoing approaches and report following metrics: Initial error(I. err) final translation error(T. err) and trajectory length(Tj. len) in meters, rotation error(R. err) in degrees. It can be observed that existing approaches are able to converge only on the simple scenes, whereas our approaches successfully converge on all. NC stands for not converging.